\documentclass[10pt,twocolumn,letterpaper]{article}

\usepackage[pagenumbers]{cvpr} 

\usepackage{graphicx}
\usepackage{amsmath}
\usepackage{amssymb}
\usepackage{booktabs}
\usepackage{bm}
\usepackage[ruled, lined, linesnumbered, commentsnumbered, longend]{algorithm2e}

\usepackage[pagebackref,breaklinks,colorlinks]{hyperref}

\usepackage[capitalize]{cleveref}
\crefname{section}{Sec.}{Secs.}
\Crefname{section}{Section}{Sections}
\Crefname{table}{Table}{Tables}
\crefname{table}{Tab.}{Tabs.}

\def\cvprPaperID{*****} 
\def\confName{CVPR}
\def\confYear{2022}

\begin{document}


\title{MIS-FM: 3D Medical Image Segmentation using Foundation Models Pretrained on a Large-Scale Unannotated Dataset}

\author{Guotai Wang$^{1, 2}$, Jianghao Wu$^{1, 2}$, Xiangde Luo$^{1, 2}$, Xinglong Liu$^{3}$, Kang Li$^{4, 2}$, 
Shaoting Zhang$^{1, 2}$\\
$^1$School of Mechanical and Electrical Engineering, University of Electronic Science and Technology of China\\
$^2$Shanghai Artificial Intellgence Laboratory\\
$^3$SenseTime Research, Beijing, China \\
$^4$West China Biomedical Big Data Center, Sichuan University\\
{\tt\small guotai.wang@uestc.edu.cn}
}
\maketitle

\begin{abstract}
Pretraining with large-scale 3D volumes has a potential for improving the segmentation performance on a target medical image dataset where the training images and annotations are limited. Due to the high cost of acquiring pixel-level segmentation annotations on the large-scale pretraining dataset, pretraining with unannotated images is highly desirable. In this work, we propose a novel self-supervised learning strategy named Volume Fusion (VF) for pretraining 3D segmentation models. It fuses several random patches from a foreground sub-volume to a background sub-volume based on a predefined set of discrete fusion coefficients, and forces the model to predict the fusion coefficient of each voxel, which is formulated as a self-supervised segmentation task without manual annotations. Additionally, we propose a novel network architecture based on parallel convolution and transformer blocks that is suitable to be transferred to different downstream segmentation tasks with various scales of organs and lesions. The proposed model was pretrained with 110k unannotated 3D CT volumes, and experiments with different downstream segmentation targets including head and neck organs, thoracic/abdominal organs  showed that our pretrained model largely outperformed training from scratch and several state-of-the-art self-supervised training methods and segmentation models.  The code and pretrained model are available at \url{https://github.com/openmedlab/MIS-FM}.  
\end{abstract}

\section{Introduction}
\label{sec:intro}
Automatic segmentation of 3D medical images plays an important role in computer-assisted diagnosis and treatment planning~\cite{Shen2017}. In recent years, deep learning models such as Convolution Neural Networks (CNNs)~\cite{Isensee2021} and Transformers~\cite{He2023} have achieved state-of-the-art segmentation performance, and some of them have obtained clinically applicable results in segmentation of Organs-at-Risk (OAR) and brain lesions~\cite{Zhang2022,Shapey2019}. Their high performance heavily relies on supervised learning from a large set of training images for the given segmentation targets with pixel-level manual annotations. 
\begin{figure}[t]
	\centering
	\includegraphics[width=1.0\linewidth]{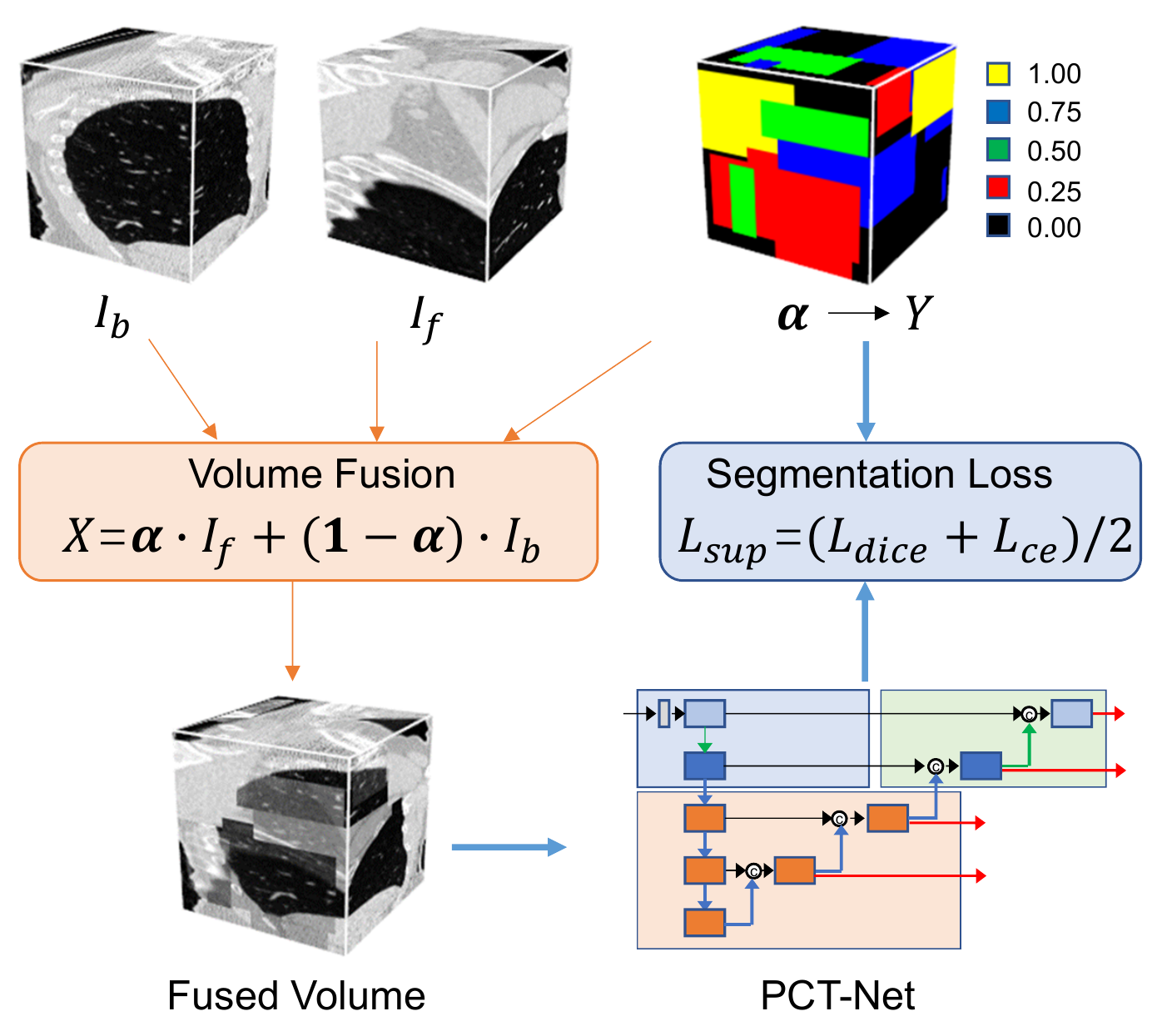}
	
	\caption{Overview of our proposed self-supervised pretraining method based on volume fusion.}
	\label{fig:overview}
\end{figure}

However, such a training paradigm is not applicable for a wide range of long-tailed diseases due to challenges related to collection of data and annotations~\cite{Tajbakhsh2019}. First, the number of images may be limited in rare diseases and non-routine imaging protocols, making it hard to collect a large set of images for training deep learning models. Second, annotating 3D medical images with pixel-level labels requires expert knowledge and is time-consuming, which makes it expensive and difficult to obtain segmentation annotations for a large training set. As a result, 
when the available training set is small, which is the case for most diseases and targets of interest, the segmentation performance of deep learning models is still limited. 

To deal with this problem, one promising solution is to pretrain the segmentation model on available large-scale datasets, such as public datasets in the community and in-house datasets at hand that are easier to obtain than images for the target segmentation task~\cite{Zhang2023c}. As pretraining with a large-scale dataset can learn diverse features that are potentially general to different downstream tasks, fine-tuning the model on a target dataset can not only improve the convergence speed, but also reduce the risk of over-fitting, leading to higher performance.
For example, models pretrained on ImageNet have been widely used for detection and segmentation models in computer vision~\cite{Chen2016deeplab,Dosovitskiy2020}. Recently, large-scale Contrastive Language-Image Pretraining (CLIP) models~\cite{Radford2021} have also showed excellent generalizability in image recognition tasks. However, these models pretrained on natural images may have a limited performance on medical images due to their differences in contrast, image styles and object categories. In addition, extracting 3D features is essential for 3D medical image segmentation, while the pretrained models for general computer vision only deal with 2D images.  Therefore, it is desirable to pretrain a model on large-scale 3D medical images to boost the performance in downstream segmentation tasks.

Recently, some researches have tried to leverage public medical image datasets with partial or full annotations to pretrain 3D segmentation models using supervised learning~\cite{Liu2023,Huang2023stu}. Despite their promising performance, the supervised pretraining paradigm is limited by the high annotation cost and not scalable to thousands of unannotated images. Instead, pretraining with Self-Supervised Learning (SSL) is more appealing when the large-scale pretraining dataset is not annotated, which is the common case in practice~\cite{Tang2022}. Most existing SSL strategies are based on image/patch-level classification or contrastive learning~\cite{Chen2020e,Caron2021,Gidaris2018a} to learn an effective feature extractor (i.e, encoder), while segmentation models usually have an additional decoder that is not used in the pretext tasks during pretraining, which may make the self-supervised pretraining less effective. Using image restoration~\cite{Zhou2021,He2022mae,Gao2021b} as the pretext task in SSL can alleviate this problem by pretraining the encoder and encoder for pixel-level intensity prediction. However, the mismatch between the pretraining and downstream tasks (i.e., regression vs segmentation) may hinder the feature transferability and limit the segmentation performance. 

To address these issues, we propose a novel self-supervised pretraining strategy named Volume Fusion (VF) for 3D medical image segmentation. Differently from existing SSL methods that define pretext tasks as image-level classification or image restoration, our method uses a pseudo-segmentation pretext task that pretrains a 3D segmentation model directly. Specifically, it fuses a foreground sub-volume from a 3D image with a background sub-volume from another image based on a fusion coefficient map, and the coefficient  of each voxel takes values from  a discrete set where each element is treated as a class. The model to pretrain takes the fused sub-volume as input, and predicts the class label of each voxel, which is exactly a supervised segmentation pretext task, as illustrated in Fig.~\ref{fig:overview}. It has several advantages over existing pretraining approaches for medical image segmentation models. First, compared with existing supervised learning-based pretraining~\cite{Liu2023,Huang2023stu}, it generates 
paired input images and segmentation label maps automatically, which does not require manual annotations, making it be able to easily leverage unannotated images at a larger scale for pretraining at zero annotation cost. Second, compared with methods that only pretrain a feature encoder~\cite{Chen2020e,Gidaris2018a,Caron2021}, it pretrains the encoder and decoder at the same time, leading the entire segmentation model to be sufficiently pretrained. In addition, compared with image restoration~\cite{Zhou2021,He2022mae,Gao2021b}, the gap between pretraining and downstream tasks are minimized, as both of them are formulated as segmentation tasks, which makes the features learned during pretraining be more transferable to downstream segmentation.   

The contribution of this work is three-fold:
\begin{itemize}
\item First, we propose Volume Fusion, a novel self-supervised learning method to pretrain 3D medical image segmentation models. It is free of manual annotation, and enables pretraining the entire segmentation model from large-scale unannotated images by using a pseudo-segmentation pretext task to match the downstream segmentation task. 
\item Second, we propose a Parallel Convolution and Transformer Network (PCT-Net) that combines local and long-range feature learning for medical image segmentation. It has a strong feature learning ability with moderate model size, which is suitable for pretraining and can be used on a common GPU device.
\item Thirdly, we release a pretrained PCT-Net based on 110k unannotated 3D Computer Tomography (CT) scans of various body parts. To the best of our knowledge, this is the largest medical image dataset for pretraining so far, while the pretraining datasets are much smaller in existing works (around 5k at most)~\cite{Tang2022}. Effectiveness of our pretrained model is validated with several downstream tasks for segmentation of head and neck, thoracic and abdominal organs, respectively.
\end{itemize}

\section{Related Works}
\label{sec:related_works}
\subsection{Medical Image Segmentation Models}
Deep learning models for medical image segmentation can be roughly divided into CNN-based, Transformer-based and hybrid models that combine CNN and Transformer. CNN-based models are dominated by U-Net-like neural networks~\cite{Ronneberger2015,Abdulkadir2016,Isensee2021} with encoder-decoder structures. Many new blocks have been introduced to extend U-Net for better performance, such as residual connection~\cite{Milletari2016}, nested connection~\cite{Zhou2018}, spatial and channel attentions~\cite{Roy2019,Gu2020a}, multi-scale feature learning~\cite{Gu2019cenet} and 2.5D convolution~\cite{Wang2022tmi}. 

Recently, Vision Transformers (ViT)~\cite{Dosovitskiy2020,Liu2021swin,He2023} based on self-attention have been increasingly investigated for segmentation. For 2D segmentation, He et al.~\cite{He2022} proposed a Fully Transformer Network (FTN) to learn long-range contextual information. MedT~\cite{Valanarasu2021a} uses a dual-branch Transformer with gated axial attention to reduce the computational cost. Swin-UNet~\cite{Cao2021} is a U-Net-like pure Transformer that uses swin Transformer based on shifted windows for improving the efficiency. To segment 3D volumes,  UNETR~\cite{Hatamizadeh2022} uses a Transformer as the encoder to learn sequence representations of the input volume that are connected to a decoder for segmentation. UNETR++~\cite{Shaker2022} introduces parallel attention modules with shared key-queries to efficiently learn enriched spatial-channel feature representations. 

To combine the advantages of CNNs and Transformers, TransUNet~\cite{Chen2021a} replaces the bottleneck of UNet with Transformers for enhancing its feature representation ability. UTNet~\cite{Gao2021c} uses a sequential combination of a residual convolutional block and a Transformer in both the encoder and decoder. CoTr~\cite{Xie2021} uses a CNN-based encoder  to extract feature representations and an efficient deformable Transformer to capture long-range dependency. 
The nnFormer~\cite{Zhou2021a} uses interleaved convolution and self-attention, where  local and global volume-based self-attention mechanism is used to learn volume representations.

\subsection{Pretrained Models for Segmentation}
In early studies, supervised training with labeled images from computer vision community has demonstrated improved performance on medical images after transfer learning~\cite{Lu2016a,Raghu2019}. To better learn features from medical images, Liu et al.~\cite{Liu2023} trained a segmentation model from 3.4k public CT scans with partial annotations based on CLIP, and demonstrated its stronger transfer learning performance on novel tasks.
Huang et al.~\cite{Huang2023stu} leveraged the TotalSegmentator dataset~\cite{Wasserthal2022} with annotations of 104 anatomical structures in 1.2k CT scans to train a scaled U-Net. However, acquiring accurate  segmentation annotations of multiple structures in 3D medical images is extremely time-consuming and expensive, making these methods hardly applicable to larger pretraining datasets. To get rid of the annotations, Model Genesis~\cite{Zhou2021} uses image restoration-based self-supervision to pretrain CNNs from 623 chest CT scans. 
 Swin UNETR~\cite{Tang2022} uses self-supervised learning to pretrain swin Transformers, making it possible to pretrain with a larger unannotated dataset of 5k  public CT scans. However, the pretraining datasets in these works are relatively small compared with ours ($\approx$110k CT scans), and the Swin UNETR only uses Transformers  in the encoder, with limited long-range feature learning ability in the decoder.

\subsection{Self-Supervised Learning}
Self-supervised learning has been widely used for visual representation learning via pretext tasks without human annotations~\cite{Jing2021}. Pretext tasks can be defined in various formulations. The first type is classification, such as prediction of discrete rotation angles~\cite{Gidaris2018a}, jigsaw puzzle solving~\cite{Gevers2016} and self-distillation~\cite{Caron2021,Ye2022a}. The second type is contrastive learning~\cite{Chen2020e,He2020b} that maximizes the similarity between positive pairs and minimizes that between negative pairs based on a contrastive loss. In addition, some image restoration methods have been proposed, such as image colorization~\cite{Zhang2016g}, denoising~\cite{Gao2021b}, inpainting~\cite{Pathak2016a} and Masked Auto-Encoder (MAE)~\cite{He2022mae}.

For 3D medical images, Chen et al.~\cite{Chen2019c} proposed context restoration with random swap of local patches. Zhu et al.~\cite{Zhu2020b} proposed Rubik's Cube+  where the pretext task involves cube ordering, cube rotating and cube masking. Zhou et al.~\cite{Zhou2021a} combined non-linear transformation, local pixel shuffling and outer/inner cutouts for image restoration. Tang et al.~\cite{Tang2022} combined inpainting, contrastive learning and rotation prediction for SSL. Chen et al.~\cite{Chen2023} proposed Masked Image Modeling (MIM) that is a 3D extension of MAE. However, the pretext task of image restoration in these works is different from the downstream segmentation task, where the gap may limit the transfer learning performance.


\section{Methods}

Fig.~\ref{fig:overview} shows an overview of our proposed method for pretraining with unannotated 3D medical images. We introduce a pretext task based on pseudo-segmentation, where Volume Fusion (VF) is proposed  to generate paired images and segmentation labels to pretrain the 3D segmentation model, which can better match the downstream task of segmentation than existing SSL methods. The pretraining strategy is combined with our proposed PCT-Net to obtain a pretrained model that is applied to segmentation of different objects from 3D medical images after fine tuning with a small set of labeled data.

\subsection{Volume Fusion for Self-Supervised Learning}

The main idea of Volume Fusion-based SSL is to fuse two sub-volumes with discrete voxel-level coefficients (i.e., different fusion categories), and train a model taking the fused sub-volume as input to predict the fusion category of each voxel, which is a supervised segmentation pretext task.     
\subsubsection{Volume Fusion}
Let $I_{b}$ and $I_{f}$ denote two sub-volumes cropped from two different 3D scans, respectively, and their size is denoted as $D\times H\times W$, where $D$, $H$ and $W$ are the depth, height and width, respectively. We use $I_{b}$ as the background sub-volume, and  $I_{f}$ as the foreground sub-volume, and they are fused into a new sub-volume $X \in \mathbb{R}^{D\times H\times W}$:

\begin{equation}
X = \bm{\alpha} I_f + (\bm{1}-\bm{\alpha}) I_b
\end{equation}
where $\bm{\alpha}$ is the fusion coefficient map, and its $i$-th element $\alpha_i$ represent the coefficient for voxel $i$. $\alpha_i$ takes values from a discrete set $\mathcal{V}$ =$ \{0.0, 1/K, 2/K, ..., (K-1) / K,  1.0 \}$, where $K$ is the number of non-zero fusion coefficients, and it corresponds to the foreground class number in the fusion output.

To model the pretext task as a segmentation task, we take each element in $\mathcal{V}$ as a class, leading to $C = K+1$ classes of fused voxels in $X$. We use $Y$ to denote the class label map corresponding to $\bm{\alpha}$, and the $i$-th voxel's class label $y_i$ is 0, 1, 2, ..., $K$ for $\alpha_i$ = 0.0, $1/K$, $2/K$, ..., 1.0, respectively. 

\begin{algorithm}
	\caption{SSL based on Volume Fusion}\label{alg}
	\KwIn{Unannotated dataset $\mathcal{D}$. Learning rate $\eta$. Batch size $B$. Foreground class number $K$. }
	\KwOut{Optimized network parameter $\theta$.}
	Randomly initialize $\theta$  \\
	\While{Stop condition not reached}{
		Initialize training batch $\mathcal{B}$ = $\emptyset$.\\
		\For{$b$ = $1$, $2$, ..., $B$}{
			$I_b$, $I_f$ $\leftarrow$ Sub-volumes cropped from two different 3D scans in $\mathcal{D}$. \\
			Initializea $\bm{\alpha}$ as zero array\\
			$M$ $\leftarrow$ Sampling from $U\sim(m_0, m_1)$\\
			\For{$m$ = $1$, $2$, ..., $M$}{
				$\mathbf{x}_m$ $\leftarrow$ A random local patch \\
				$\alpha_m$ $\leftarrow$ Sampling from \{1/\textit{K}, 2/\textit{K}, ..., 1\} \\
				$\bm{\alpha}$[$\mathbf{x}_m$] = $\alpha_m$
			}
			$X$  $\leftarrow$ $\bm{\alpha}I_f$ + $(\bm{1} - \bm{\alpha})I_b$ \\
			$Y$ $\leftarrow$ Covnert $\bm{\alpha}$ to a label map \\
			$\mathcal{B}$ $\leftarrow$ $\mathcal{B}\cup \{(X, Y)\}$ 
		}
		Forward pass with the batch $\mathcal{B}$ \\
		
		$L_{sup}$ $\leftarrow$ Calculate supervised loss using Eq.~\ref{eq:l_sup} \\
		$\theta$  $\leftarrow$  $\theta -$$\eta \nabla_{\theta}L_{sup}$
	}
\end{algorithm}

\begin{figure}[t]
	\centering
	\includegraphics[width=1.0\linewidth]{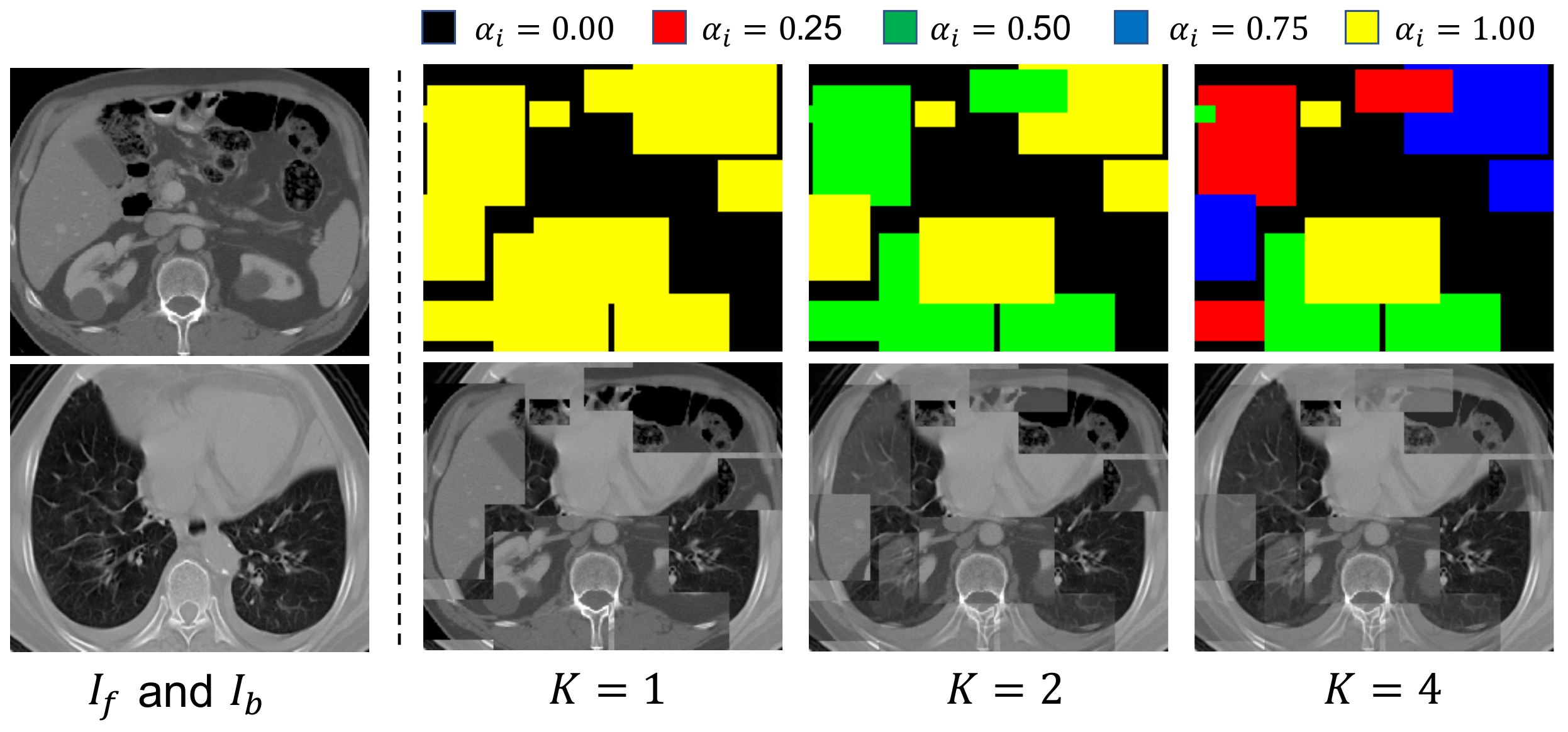}
	
	\caption{2D illustration of fused images with different $K$ values. The colorful images are fusion coefficient maps, and they are converted into label maps for the pretext segmentation task.}
	\label{fig:example1}
\end{figure}

\subsubsection{Pretext Task based on Segmentation}

During the pretraining process, for each pair of $I_b$ and $I_f$, we generate the fusion coefficient map $\bm{\alpha}$ by setting the fusion coefficient values of different local patches randomly. Specifically, assume the foreground patch number in $\bm{\alpha}$ is $M\sim U(m_0, m_1)$, where $m_0$ and $m_1$ are the minimal and maximal possible values of $M$, respectively. We select $M$ foreground patches at random sizes sequentially, and set the fusion coefficient in each patch as a random value from \{$1/K$, $2/K$, ..., 1.0\}. For voxels outside of these patches (i.e., the background region), the $\alpha_i$ value is set to 0.0. Then the generated  $\bm{\alpha}$ is used to fuse   $I_b$ and $I_f$ to obtain $X$, and the corresponding label map $Y$ converted from $\bm{\alpha}$ is used as the segmentation ground truth for $X$. Then a supervised segmentation loss $L_{sup}$ based on Dice loss $L_{dice}$ and cross entropy loss $L_{ce}$ is used to pretrain the segmentation model based on the paired $X$ and $Y$:

\begin{equation}\label{eq:l_sup}
	L_{sup} = \frac{1}{2}(L_{dice} + L_{ce})
\end{equation}
\begin{equation}
	L_{dice} = 1 - \frac{1}{CV}\sum_{c=0}^{C-1}\sum_{i=0}^{V-1}\frac{2p_i^cy_i^c}{p_i^c + y_i^c + \epsilon}
\end{equation}
\begin{equation}
	L_{ce} = - \frac{1}{V}\sum_{i=0}^{V-1}\sum_{c=0}^{C-1} y_i^c log(p_i^c)
\end{equation}
where $V=D\times H \times W$ is the voxel number. $p_i^c$ is the predicted probability
of voxel $i$ being class $c$, and $y_i^c$ is the corresponding one-hot ground truth value obtained from $Y$. The pseudo code of SSL based on Volume Fusion is shown in Algorithm~\ref{alg}.

\begin{figure*}[t]
	\centering
	\includegraphics[width=0.85\linewidth]{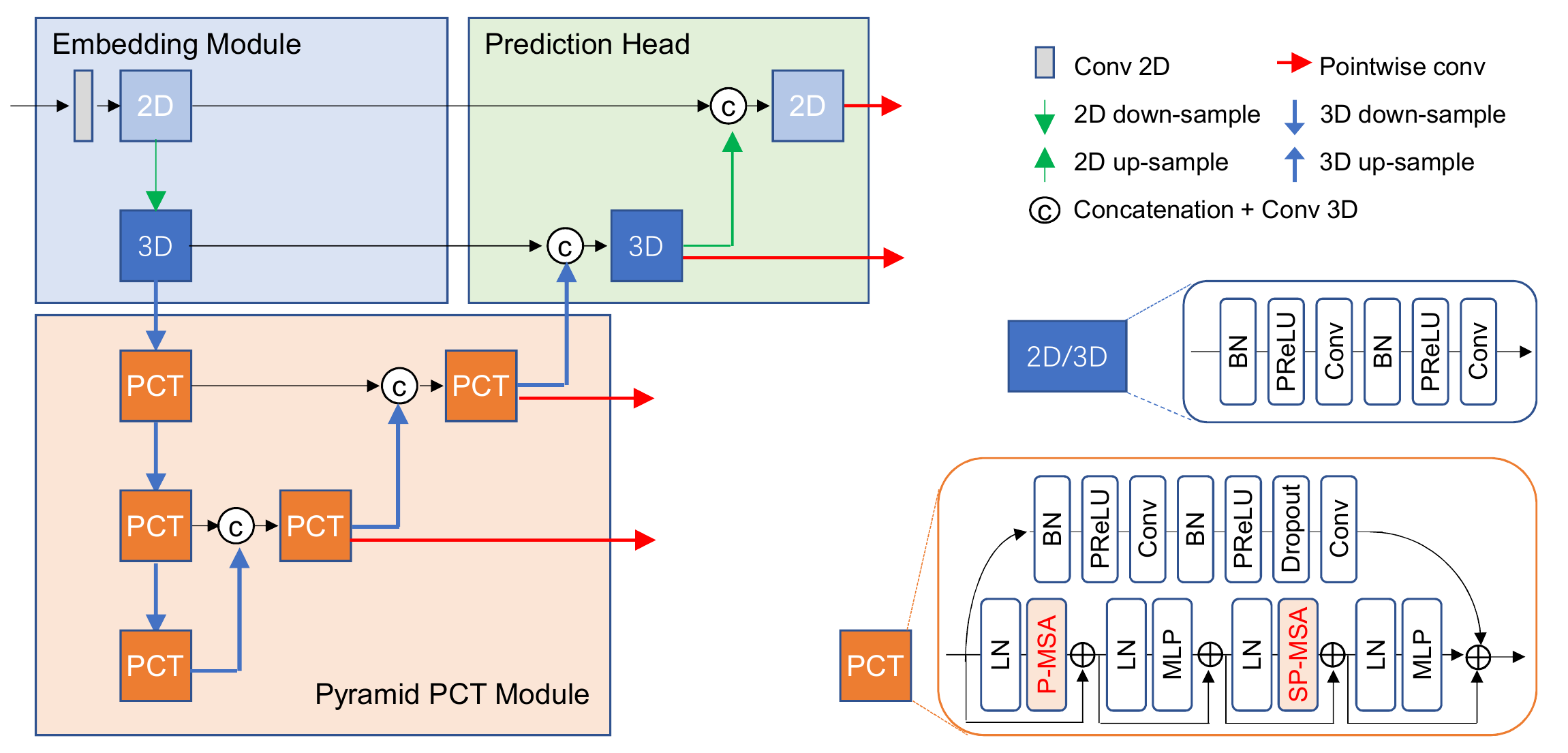}
	
	\caption{Structure of our proposed PCT-Net.}
	\label{fig:network}
\end{figure*}

\subsubsection{Context Learning Ability of Volume Fusion}
The effectiveness of pretraining is largely dependent on the feature representations learned in the pretext task, and the ability to recognize contextual semantics is essential for medical image segmentation~\cite{Chen2019c,Zhou2021}. Therefore, a good pretraining strategy for segmentation models should have high context learning ability. In SSL based on Volume Fusion, the model is trained to recognize the fusion category of each voxel. For a given fused voxel, as there are numerous possible combinations of the original foreground and background intensities, one cannot directly find a mapping from a fused voxel to the fusion category. Successful recognition of the voxel-level fusion category requires being aware of the global context of the foreground and background sub-volumes.

Fig.~\ref{fig:example1} shows an example of fused images with different $K$ values. In the case of $K$ = 1, the model is trained for a binary segmentation of the foreground patches from the background region. As both of the foreground and background sub-volumes are  CT scans with similar intensity distributions, the model needs to find the structural discontinuity around the border of patches for the segmentation, which forces it to have a context learning ability. Note that at the center of a large patch, the structural discontinuity cannot be found based on local semantics, and the model needs to be aware of the global context to successfully recognize the category.  When $K>1$, there are more types of foreground patches with soft fusion coefficients, and the model needs a higher contextual semantics recognition ability for the segmentation task. Note that the foreground patches are randomly and sequentially  generated at various scales and aspect ratios, and the region for each class can have irregular shapes due to  overlapping of the sequential patches. Therefore, the model is encouraged to learn semantics at multiple scales in our pretext task, which is highly transferable to the downstream segmentation task for objects at various scales.

\subsubsection{PCT-Net}
For medical image segmentation, CNNs are good at learning local features due to the local convolution operations. Transformers based on self-attention can better capture long-range dependency and global context. However, the patch embedding operation in Transformers  often reduces the image resolution largely~\cite{Dosovitskiy2020,Zhou2021a} (i.e., by at least 4 times), which has a limited performance on tiny structures in medical images. In addition, the self-attention operation and Multi-Layer Perceptron (MLP) in Transformers are computationally expensive. Therefore, it is desirable to combine CNNs and Transformers to design a 3D segmentation model that has both local and global feature learning ability with a moderate model size and computational cost.  

The structure of our proposed PCT-Net is illustrated in Fig.~\ref{fig:network}. It has three main components: a multi-scale feature embedding module, a pyramid Parallel Convolution and Transformer (PCT) module, and a prediction head. The embedding module uses convolutional layers to extract local features at two resolution levels. Considering that medical images usually have an anisotropic 3D resolution where the intra-plane pixel spacing is lower than inter-plane spacing~\cite{Wang2022tmi,Zhou2021a}, we use a 2D convolutional block at the highest resolution level, which is followed by a 2D downsampling. After the 2D downsampling, the 3D resolution is less anisotropic, and a 3D convolution block is used in the second resolution level. Each 2D/3D convolution block  consists of two basic convolutional layers, with each containing a Batch Normalization (BN), a Parametric Rectified Linear Unit (PReLU) activation and a 2D/3D convolution operation. In addition, before the 2D convolution block, a project layer based on 2D convolution is used to convert the input image to a high-dimension feature map. 

The pyramid PCT module has three resolution levels with an encoder-decoder structure equipped with PCT blocks. A PCT block has two branches.  The first branch uses convolutional layers to extract local features. It has a structure similar to the 3D convolution block used in the embedding module, and the difference is that a dropout layer is used before the second 3D convolution operation. The second branch consists of two self-attention blocks to learn long-range dependency. Considering the large voxel number in 3D volumes, we use a 3D version of Swin Transformer~\cite{Liu2021swin} to reduce the cost of computation and memory, i.e., Patch-based Multi-head Self-Attention (P-MSA) that applies self attention at each of non-overlapping patches instead of the entire volume .  The first self-attention block has a Layer Norm (LN), a P-MSA, followed by another LN and a MLP. The second self-attention block is similar to the first one, and it uses a shifted version of the P-MSA (SP-MSA), where the shift size is half of the patch size. The outputs of these two branches in a PCT block are added together so that the complementary  local and global features can be leveraged. 

The prediction head recovers high-resolution feature maps that is symmetric to the embedding module. At each resolution level, a concatenation with convolution is used to fuse the feature from the embedding module and that upsampled from a lower resolution level, and it is followed by a 2D/3D convolutional block. We use a pointwise convolution  to predict a segmentation map at each of the two resolution levels. In addition, two pointwise convolutions are used at the first two resolution levels of the pyramid PCT module respectively, so that PCT-Net generates predictions at four different scales that are used for deep supervision.

\section{Experiments and Results}
\subsection{Datasets}
\textbf{Pretraining Datasets:} We collected 3D CT scans from several public datasets and a private dataset to construct the pretraining images at  three different magnitudes: PData-1k, PData-10k and PData-110k where the 3D image number was around 1k, 10k and 110k respectively.

PData-1k contains  1,065 public CT scans, where 298 head and neck CT scans are from the public TCIA Head-Neck-PET-CT dataset\footnote{https://wiki.cancerimagingarchive.net/display/Public/Head-Neck-PET-CT}, 267 lung CT scans are from the public LUNA dataset\footnote{https://luna16.grand-challenge.org} (subset 0-2), and 500 abdominal CT scans are from the public FLARE22 dataset\footnote{https://flare22.grand-challenge.org/}. 

PData-10k contains  10,137 public CT scans each from a different patient: 2,037 head and neck CT scans are from  TCIA Head-Neck-PET-CT, OPC-Radiomics\footnote{https://wiki.cancerimagingarchive.net/pages/viewpage.action?pageId=33948764}, HNSCC\footnote{https://wiki.cancerimagingarchive.net/display/Public/HNSCC}, QIN-HEADNECK\footnote{https://wiki.cancerimagingarchive.net/display/Public/QIN-HEADNECK} and  TCGA-HNSC\footnote{https://wiki.cancerimagingarchive.net/pages/viewpage.action?pageId=11829589};  2,659 lung CT scans are from LUNA, TCIA Covid19\footnote{https://wiki.cancerimagingarchive.net/display/Public/CT+Images+in+COVID-19}
and LIDC-IDRI\footnote{https://wiki.cancerimagingarchive.net/pages/viewpage.action?pageId=1966254};  5,441 abdominal or whole-body CT scans are from FLARE22,  ACRIN 6664\footnote{https://wiki.cancerimagingarchive.net/pages/viewpage.action?pageId=3539213}, AbdomenCT-1K~\cite{Ma2022} and TotalSegmentator~\cite{Wasserthal2022}.

\begin{table}
	\centering
	\caption{ Details of the 3D datasets used for pretraining.}
	\resizebox{\linewidth}{!}{
		\begin{tabular}{llll}
			\toprule
			Body part & Name &   Pathology & Cases \\
			\midrule
			Head-Neck  & Head-Neck-PET-CT & Head and neck cancer & 298 \\
			 & OPC-Radiomics & Oropharynx cancer & 606 \\
			 & HNSCC   & Head and neck cancer & 627 \\
			& QIN-HeadNeck & Head and neck cancer & 279 \\
			& TCGA-HNSC & Head and neck cancer & 227 \\
			\midrule
			Chest  & LUNA & Lung nodule & 888 \\
			  & TCIA Covid19 & Covid19 & 753 \\
			  & LIDC-IDRI & Lung nodule / cancer & 1,018 \\
			\midrule
			Abdomen  & FLARE22 & Abdominal lesions & 2,300 \\
			& AbdomenCT-1K & Abdominal lesions & 1,112 \\
			& ACRIN 6664 & Colon cancer & 825 \\
			\midrule
			Whole body & TotalSegmentator & Various pathologies & 1,204 \\
			\midrule
		Chest & Private & Lung diseases & 103k \\
			\bottomrule
		\end{tabular}%
	}
	\label{tab:data}%
\end{table}%
 
 PData-110k contains all the images in PData-10k and additional private set of 103k lung CT scans, as shown in Table~\ref{tab:data}. The private dataset was collected from nine hospitals with different vendors and scanning sequences including non-contrast CT, contrast-enhanced CT and CT pulmonary angiography. The patient age ranged from 20 to 90 years. The slice thickness was 0.6$\sim$2.5 mm in 97\% images, and $\ge$5.0~mm in the other 3\% images. Around 80\% of the images have pathologies including lung nodules, pneumonedema, pneumonia, emphysema,  pleural effusion, lymphadenopathy, rib fracture and liver lesions, etc.
 
\textbf{Downstream Segmentation Datasets:} We considered three different downstream tasks for segmentation of head and neck organs, thoracic organs and abdominal organs from CT scans, respectively. The datasets are: 

1) MICCAI 2015 Head-Neck dataset~\cite{Raudaschl2017} with 48 CT scans for segmentation of Organs-at-Risks (OAR) in radiotherapy planning. The OARs include brain stem, optic chiasm, mandible, bilateral optic nerves, bilateral parotid glands and bilateral Sub-Mandibular (SM) glands. For each bilateral structure, we merged the left and right instances as a single class, leading to 6 OARs for segmentation. Each volume has 76$\sim$360 slices with a size of 512$\times$512. The in-plane resolution ranged from 0.76 to 1.27~mm, and the inter-slice spacing was 1.25$\sim$3.0~mm. The images were randomly split into 30, 6 and 12 for training, validation and testing, respectively. 

2) Segmentation of THoracic Organs-at-Risk  dataset (SegTHOR)~\cite{Lambert2020a}. The OARs include heart, aorta, trachea and esophagus. The dataset consists of 60 CT scans with an in-plane resolution of 0.90$\sim$1.37 mm and an inter-slice spacing of 2.0$\sim$3.7 mm. Each volume has 150$\sim$284 slices, with a slice size of 512$\times$512. As the annotations for the official testing set (20 cases) are not publicly available, we used the remaining 40 cases for experiments, and they were randomly split into 24, 6 and 10 for training, validation and testing, respectively.

3) Synapse for multi-organ segmentation that includes 30 cases of abdominal CT scans. Following existing works~\cite{Chen2021a,Zhou2021a}, we deal with segmentation of 8 organs, i.e., aorta, gallbladder, left kidney, right kidney, liver, pancreas, spleen and stomach. Each CT volume has 85$\sim$198 slices with a size of 512$\times$512. The in-plane resolution was 0.54$\sim$0.98 mm, and the inter-slice spacing was 2.5$\sim$5.0 mm. The 3D images were randomly split into 18, 4 and 8 for training, validation and testing, respectively.

\subsection{Implementation Details} 
Our PCT-Net and Volume Fusion were implemented in PyTorch\footnote{https://pytorch.org} and PyMIC\footnote{https://github.com/HiLab-git/PyMIC}~\cite{Wang2023a}. In PCT-Net, the convolutional kernel size was 3$\times$3 and 3$\times$3$\times$3 for the 2D and 3D convolutions, respectively. The channel numbers at the five resolution levels were 24, 48, 128, 256 and 512 respectively. For Volume Fusion-based pretraining, the sub-volume size was 64$\times$128$\times$128. To generate random fusion coefficient map $\bm{\alpha}$, the foreground class number was set as $K=4$, and the patch number was $M\sim U(10, 40)$. The ranges of depth, height and width of the local patches were 8$\sim$40,  8$\sim$80  and 8$\sim$80, respectively. 

For pretraining, we used a batch size of 4 with the Adam optimizer, weight decay of $10^{-5}$ and initial learning rate of $10^{-3}$. For PData-1k, PData-10k and PData-110k, the maximal training iteration number was 100k, 200k and 400k, respectively, and the learning rate was halved for every 20k, 40k and 50k, respectively. For pretraining with PData-110k, a server with two NVIDIA Tesla A100 GPUs (80 GB memory) was used, while the other pretraining and downstream training/inference procedures were conducted on a Ubuntu desktop that has two NVIDIA GTX 3090  GPUs (24 GB memory). The combination of Dice loss and cross entropy loss was used for both the pretraining and downstream training tasks. For a downstream task, the model was trained for 30k with a batch size of 2, and we set the initial learning rate to $10^{-3}$ that was halved at every 5k iterations. For quantitative evaluation, we used the Dice similarity coefficient (Dice) and Average Symmetric Surface Distance (ASSD) between segmentation results and the corresponding ground truths.
 
 \begin{figure}[t]
 	\centering
 	\includegraphics[width=1.0\linewidth]{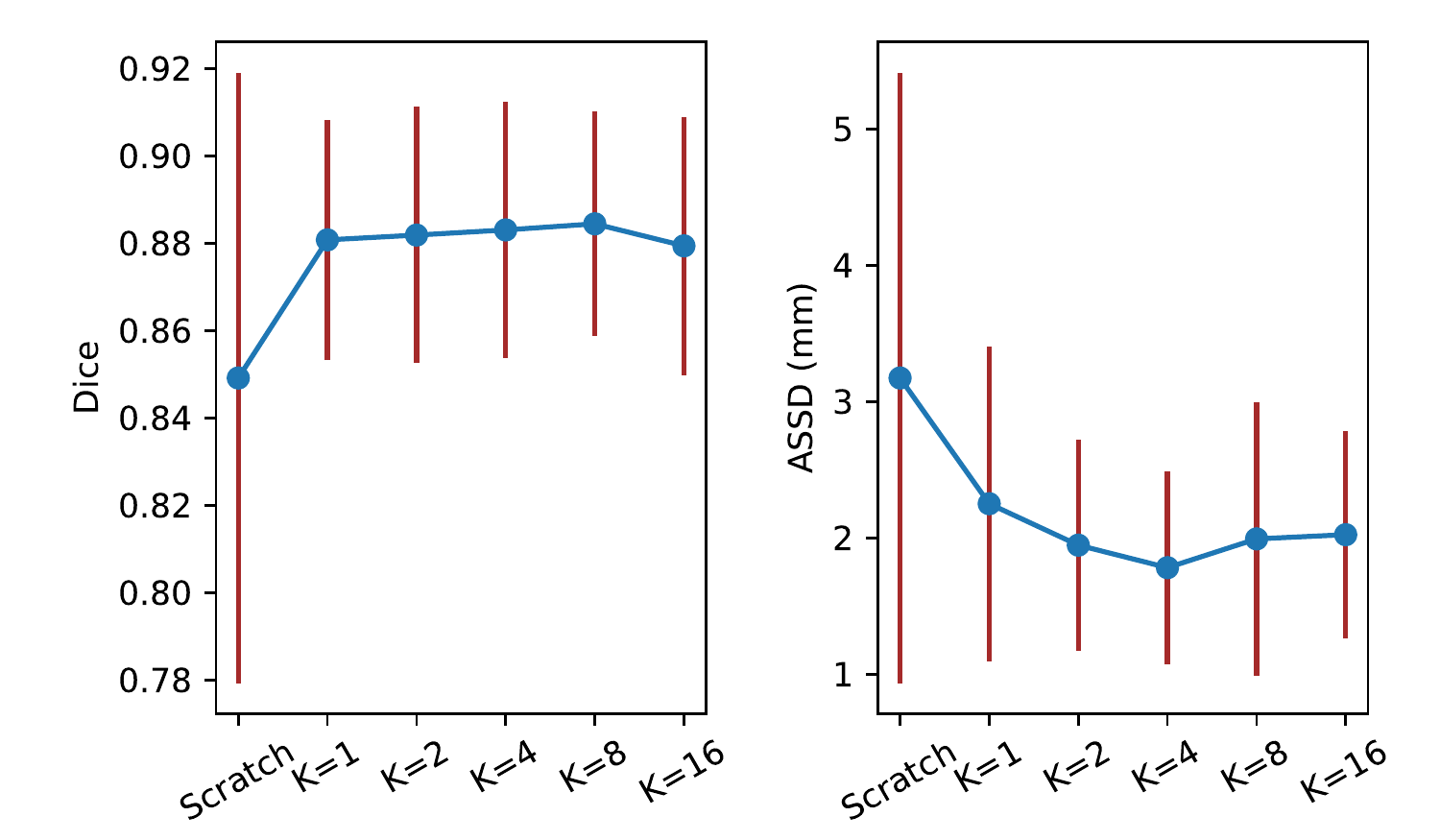}
 	
 	\caption{Performance of Volume Fusion with different $K$ values. 3D U-Net  was used as the segmentation model, and it was pretrained on PData-1K and then transferred to SegTHOR dataset.}
 	\label{fig:k_value}
 \end{figure}

\begin{figure}[t]
	\centering
	\includegraphics[width=1.0\linewidth]{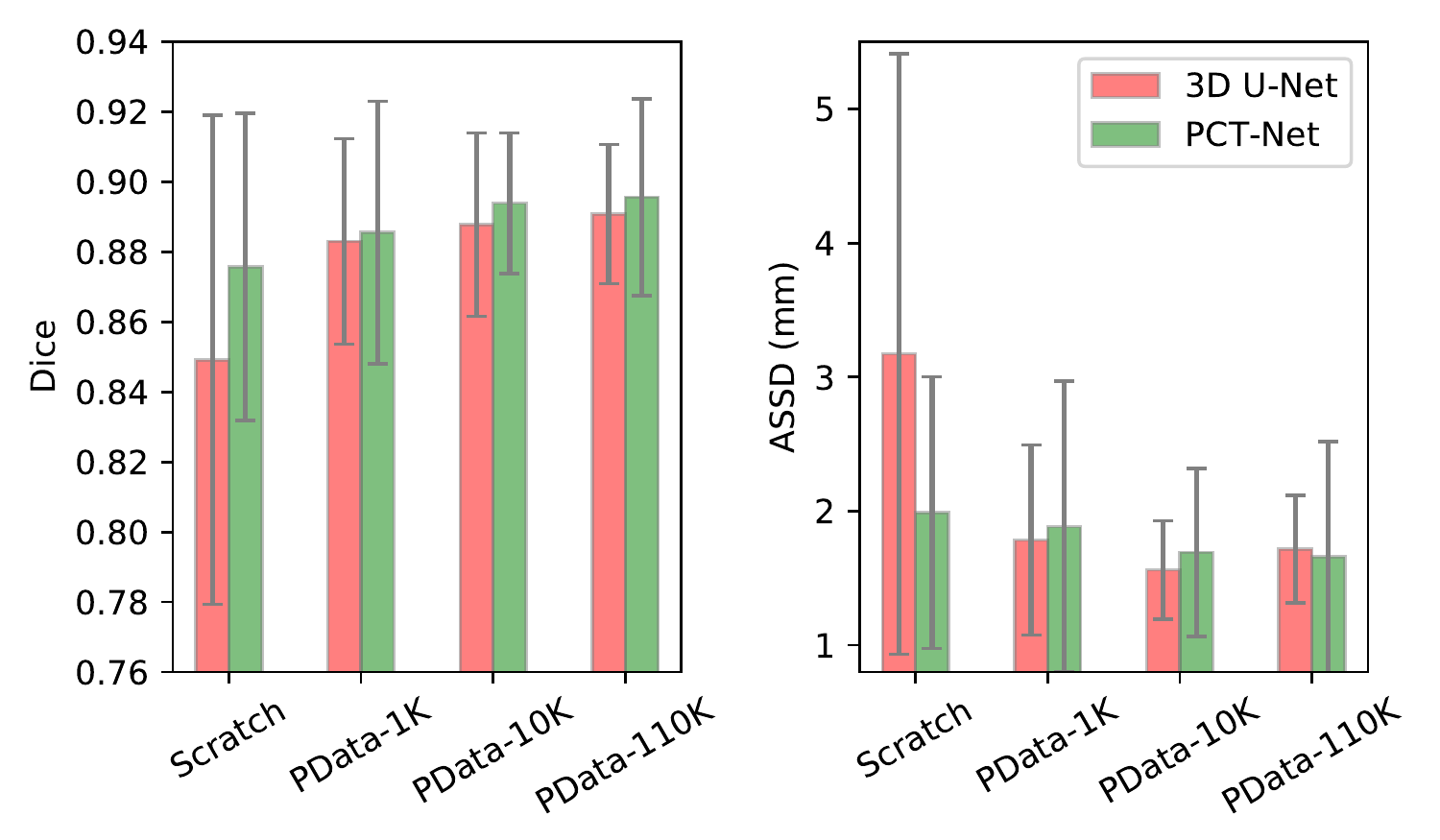}
	
	\caption{Effect of the scale of pretraining data used by  Volume Fusion on the segmentation performance on  SegTHOR dataset.}
	\label{fig:data_scale}
\end{figure}

\begin{table*}
	\centering
	\caption{ Quantitative comparison between models with different pretraining strategies on the SegTHOR dataset. PData-1K was used as the pretraining dataset, and the model architecture was 3D U-Net~\cite{Abdulkadir2016}.}
	\resizebox{\linewidth}{!}
	{
		\begin{tabular}{l|lllll|lllll}
			\toprule
			Method & \multicolumn{5}{c|}{Dice (\%)} & \multicolumn{5}{c}{ASSD (mm)}  \\
			\cline{2-11}
			& Esophagus &   Heart & Trachea & Aorta & Average &  Esophagus &   Heart & Trachea & Aorta & Average \\
			\midrule
			Scratch  & 71.87$\pm$11.27 & 90.82$\pm$4.72  & 87.63$\pm$5.77 & 89.34$\pm$6.19 & 84.92$\pm$6.47 & 3.37$\pm$2.48 & 5.52$\pm$3.98& 1.51$\pm$0.69 & 2.30$\pm$1.80 & 3.17$\pm$2.24 \\
			Patch Swapping~\cite{Chen2019c}&   75.91$\pm$9.04 & 91.88$\pm$5.27 & 87.31$\pm$6.45 & 89.64$\pm$7.60 & 86.18$\pm$4.05 & 2.05$\pm$0.92  & 2.96$\pm$1.64  & 3.09$\pm$5.75 & 3.02$\pm$3.41 & 2.78$\pm$1.54 \\
			Model Genesis~\cite{Zhou2021}&   76.98$\pm$8.29 & 92.78$\pm$3.13   & 87.87$\pm$6.28  & 89.64$\pm$7.74 &  86.81$\pm$4.00 & 2.15$\pm$1.00  & \textbf{2.81$\pm$1.43}  & 
			2.86$\pm$6.00  & 3.02$\pm$3.26 & 2.71$\pm$1.58 \\
			MIM~\cite{Chen2023}& 76.29$\pm$8.91  & 91.85$\pm$2.82  & 87.52$\pm$5.43  & 92.19$\pm$ 2.98 & 86.97$\pm$3.50 & 2.05$\pm$ 0.77 & 4.65$\pm$ 2.25 & 3.20$\pm$4.61& 2.22$\pm$ 1.66 & 3.03$\pm$1.55\\
			Volume Fusion & \textbf{77.61$\pm$7.82}  & \textbf{93.72$\pm$2.28}   & \textbf{88.21$\pm$4.18}   & \textbf{93.67$\pm$1.68} & \textbf{88.30$\pm$2.93} & \textbf{1.84$\pm$0.76} & 2.90$\pm$2.55 & \textbf{1.30$\pm$0.57} & \textbf{1.08$\pm$0.48} & \textbf{1.78$\pm$0.71} \\
			\bottomrule
		\end{tabular}%
	}
	\label{tab:ssl_method}%
\end{table*}%

\begin{figure*}[t]
	\centering
	\includegraphics[width=1.0\linewidth]{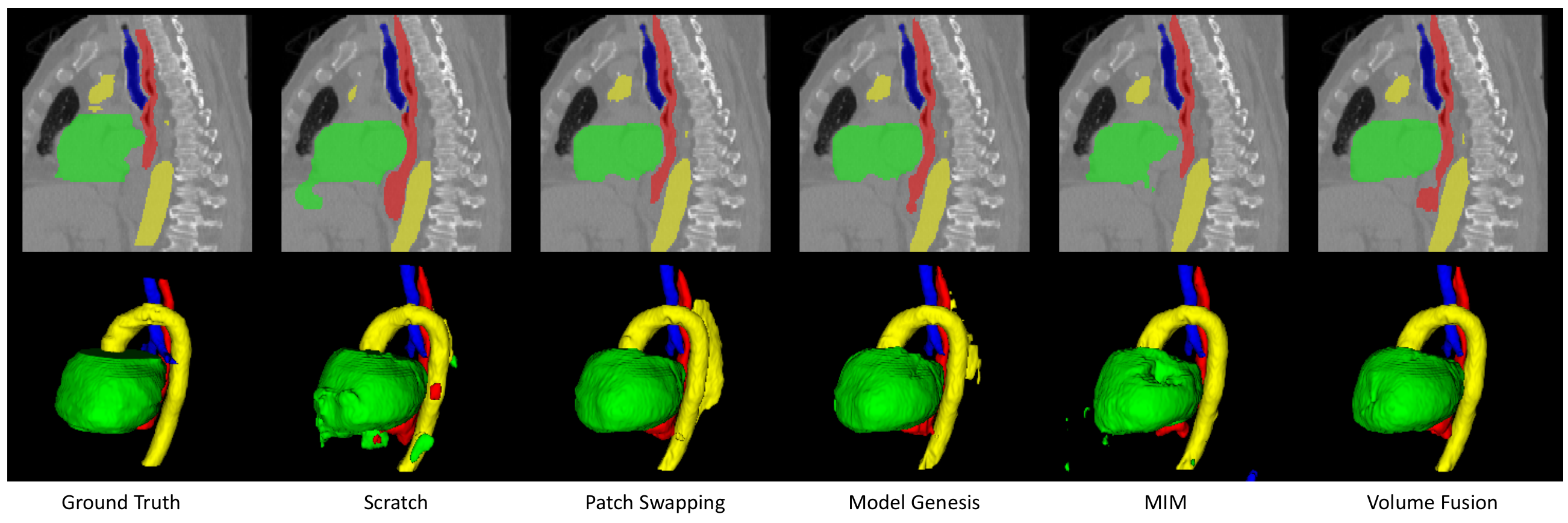}
	
	\caption{Visual comparison between results on SegTHOR dataset obtained by models pretrained with different SSL methods.}
	\label{fig:segthor_ssl}
\end{figure*}

\subsection{Ablation Study}
In this section, we conduct ablation studies to investigate the performance of our Volume Fusion with different $K$ values and different scales of pretraining data, where 3D UNet~\cite{Abdulkadir2016} and SegTHOR were employed as the segmentation model and downstream segmentation dataset, respectively.

\textbf{Effect of $K$.} To observe the effect of different $K$ values on performance of the pretrained model, we set it to 1, 2, 4, 8 and 16 respectively, and pretrained 3D U-Net~\cite{Abdulkadir2016} on PData-1K before training with the SegTHOR dataset. The pretrained models were compared with training from scratch. Quantitative evaluation results on the SegTHOR testing images are shown in Fig.~\ref{fig:k_value}. It can be observed that the average Dice value obtained by training from scratch was 84.92\%, and all the variants of Volume Fusion outperformed it. Especially, $K=1$ led to an average Dice of 88.08\%, and increasing  $K$ to 2, 4 and 8 further improved the performance, respectively. $K=16$ obtained an average Dice of 87.94\%, which was lower than those of the other $K$ values, but it was still much better than learning from scratch. In addition,  Fig.~\ref{fig:k_value} shows that $K=4$ achieved the lowest average ASSD value of 1.78~mm, compared with 3.17~mm obtained by training from scratch.

\textbf{Effect of the Scale of Pretraining data.} We further compared pretraining with PData-1K, PData-10K and PData-110K using 3D U-Net and PCT-Net, respectively. The results on the SegTHOR dataset when $K=4$ are shown in Fig.~\ref{fig:data_scale}. For 3D U-Net, PData-1K, PData-10K and PData-110K improved the average Dice from 84.92\% to 88.30\%,  88.78\% and 89.09\% respectively. For PCT-Net, training from scratch obtained an average Dice of 87.57\%, and PData-1K, PData-10K and PData-110K improved it to 88.56\%,  89.39\% and 89.56\% respectively. The results show that Volume Fusion works on different network structures, and its performance can be improved by leveraging a larger unannotated pretraining dataset. 

\begin{table*}
	\centering
	\caption{ Quantitative comparison between different methods for head and neck OAR segmentation on MICCAI 2015 Head-Neck dataset.}
	\resizebox{\linewidth}{!}
	{
		\begin{tabular}{l|l|lllllll}
			\toprule
			& Method &  Brain stem &   Optic chiasm & Mandible & Optic nerves & Parotid Glands &  SM glands &  Average \\
			\midrule
			Dice (\%) & nnU-Net~\cite{Isensee2021}  & 89.27$\pm$2.42  & 57.47$\pm$24.24  & 90.12$\pm$5.84 & 72.50$\pm$7.85 & \textbf{87.53$\pm$3.43} & 75.06$\pm$12.58 & 78.66$\pm$5.42 \\
			& TransUNet~\cite{Chen2021a}& 75.52$\pm$5.58   & 41.92$\pm$15.99 & 92.28$\pm$1.45 & 58.36$\pm$6.91 & 76.70$\pm$6.85 & 69.80$\pm$8.27  & 69.10$\pm$3.08  \\
			& nnFormer~\cite{Zhou2021a}  & 80.02$\pm$3.53   & 52.72$\pm$14.70   & 87.96$\pm$2.27  & 57.34$\pm$7.72 &  75.31$\pm$7.24 & 68.25$\pm$5.53   & 70.27$\pm$4.19 \\
			& UNETR++~\cite{Shaker2022} & 87.26$\pm$2.13  &60.44$\pm$22.49  & 93.99$\pm$1.30  & 75.19$\pm$ 5.85 & 84.61$\pm$3.89 & 80.74$\pm$ 4.53 & 80.37$\pm$3.94 \\
			& PCT-Net & 89.25$\pm$1.86   & 58.09$\pm$18.59   & 94.17$\pm$1.66   & 77.04$\pm$4.84 & 87.44$\pm$3.37 & 82.49$\pm$4.55 & 81.41$\pm$3.67 \\
			& PCT-Net + VF &  \textbf{90.24$\pm$1.78}   & \textbf{62.93$\pm$20.73}   & \textbf{94.85$\pm$1.36}   & \textbf{78.11$\pm$4.04} &  87.07$\pm$3.69 & \textbf{83.25$\pm$3.90} & \textbf{82.74$\pm$3.95}  \\
			\midrule
			ASSD (mm) & nnU-Net~\cite{Isensee2021}  & 1.08$\pm$0.30  & 1.04$\pm$0.84  & 0.63$\pm$0.42 & 0.52$\pm$0.26 & 1.15$\pm$0.43 & 3.51$\pm$5.71 & 1.32$\pm$1.04 \\
			& TransUNet~\cite{Chen2021a}& 2.00$\pm$0.40  & 4.11$\pm$2.66 & 1.46$\pm$1.13 & 2.58$\pm$0.76 & 2.65$\pm$1.06 & 4.10$\pm$1.81  & 2.82$\pm$0.76  \\
			& nnFormer~\cite{Zhou2021a}  & 2.01$\pm$0.46   & 1.28$\pm$0.55   & 1.24$\pm$1.37  & 1.16$\pm$0.45 &  2.52$\pm$1.03 & 2.26$\pm$0.82   & 1.75$\pm$0.43 \\
			& UNETR++~\cite{Shaker2022} & 1.22$\pm$0.31   & 0.97$\pm$0.89 & 0.38$\pm$0.10  & 0.44$\pm$ 0.16 & 1.34$\pm$0.38 & 1.17$\pm$0.26 & 0.92$\pm$0.19 \\
			& PCT-Net &   1.14$\pm$0.29 & 1.00$\pm$0.70   & 0.41$\pm$0.17   & 0.40$\pm$0.14 & 1.14$\pm$0.50 & 1.67$\pm$2.03 & 0.96$\pm$0.33 \\
			& PCT-Net + VF & \textbf{0.92$\pm$0.26}   & \textbf{0.87$\pm$0.72}   & \textbf{0.32$\pm$0.08}   & \textbf{0.38$\pm$0.12} & \textbf{1.09$\pm$0.29} & \textbf{1.03$\pm$0.26} & \textbf{0.77$\pm$0.16}  \\
			\bottomrule
		\end{tabular}%
	}
	\label{tab:sota_hn}%
\end{table*}%

\begin{figure*}[t]
	\centering
	\includegraphics[width=1.0\linewidth]{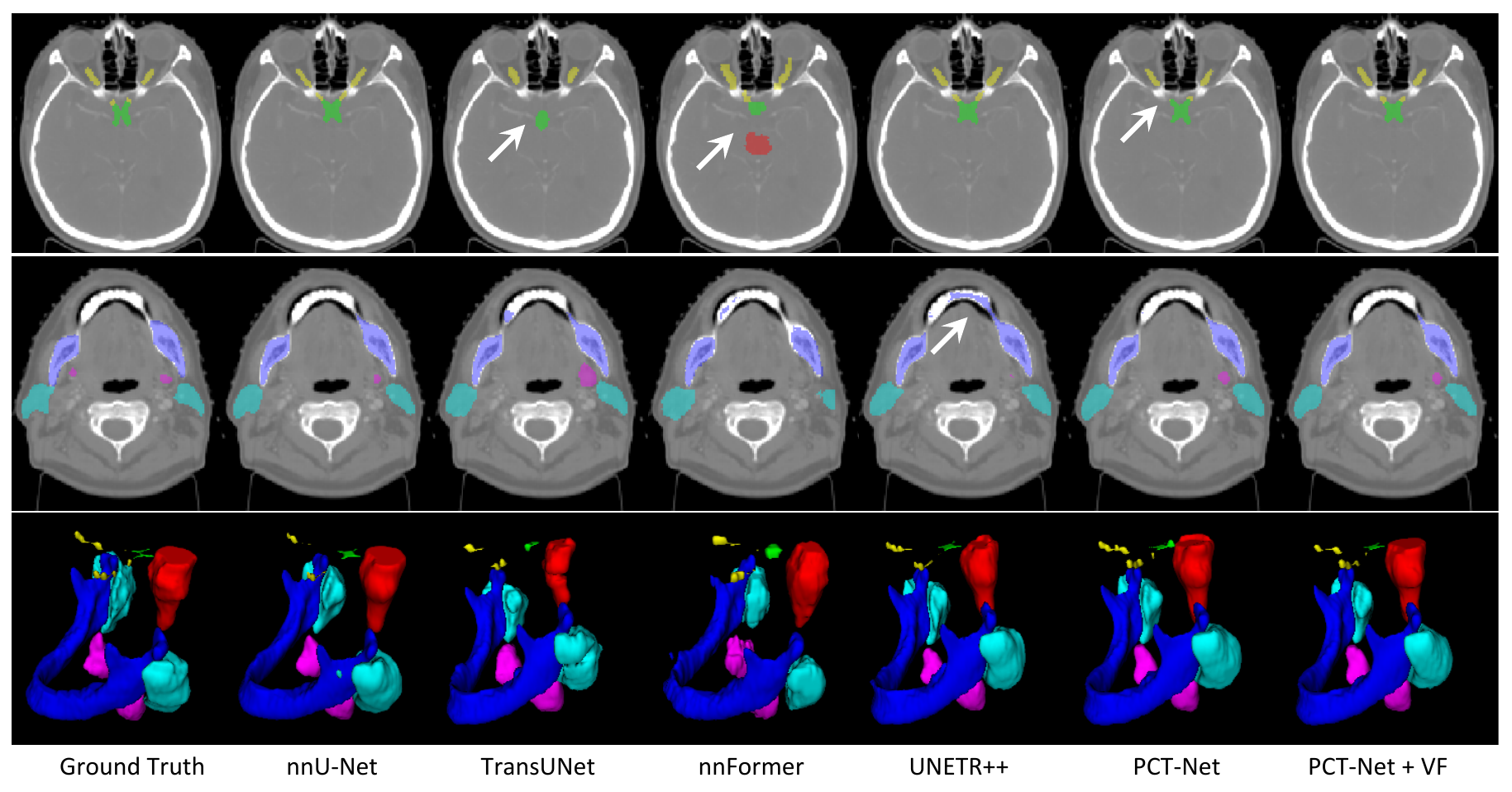}
	
	\caption{Visual comparison between results of different methods on MICCAI 2015 Head-Neck dataset. The first two rows show results in two different axial slices, and the last row presents 3D visualization. White arrows highlight the mis-segmented regions.}
	\label{fig:oar_miccai15}
\end{figure*}

\subsection{Comparison with State-of-the-art Methods}
\textbf{Comparison of Different Pretraining Methods.} 
Our proposed Volume Fusion was compared with three existing self-supervised pretraining methods for medical image segmentation models: 1) Patch Swapping~\cite{Chen2019c} where random small patches in an image are swapped for context restoration, 2) Model Genesis~\cite{Zhou2021} that combines non-linear transformation, local pixel shuffling and inter/outer cutout for image restoration, and 3) Masked Image Modeling (MIM)~\cite{Chen2023} where local patches are randomly masked at a high masking ratio for image restoration.

In this experiment, we used PData-1K to pretrain 3D UNet~\cite{Abdulkadir2016} with these different SSL methods respectively, and applied the pretrained models to the SegTHOR dataset. The quantitative evaluation results are shown in Table~\ref{tab:ssl_method}. Compared with learning from scratch, Patch Swapping~\cite{Chen2019c}, Model Genesis~\cite{Zhou2021} and MIM~\cite{Chen2023} improved the average Dice from 84.92\% to 86.18\%, 86.81\% and 86.97\%, respectively, and reduced the average ASSD value from 3.17~mm to 2.78~mm, 2.71~mm and 3.03~mm, respectively. Our proposed Volume Fusion outperformed these three methods, with an average Dice and ASSD of 88.30\% and 1.78~mm, respectively. Fig.~\ref{fig:segthor_ssl} shows a visual comparison between these methods. It can be observed that learning from scratch had an over-segmentation of the heart and esophagus, with an under-segmentation of the aorta. Patch Swapping~\cite{Chen2019c} and Model Genesis~\cite{Zhou2021} obtained an over-segmentation of the aorta, while MIM~\cite{Chen2023} obtained a poor segmentation of the heart. In contrast, Volume Fusion obtained smoother results with less mis-segmentation than the other methods.

\begin{table*}
	\centering
	\caption{ Quantitative comparison between different methods for thoracic OAR segmentation on SegTHOR dataset.}
	\resizebox{\linewidth}{!}
	{
		\begin{tabular}{l|lllll|lllll}
			\toprule
			Method & \multicolumn{5}{c|}{Dice (\%)} & \multicolumn{5}{c}{ASSD (mm)}  \\
			\cline{2-11}
			& Esophagus &   Heart & Trachea & Aorta & Average &  Esophagus &   Heart & Trachea & Aorta & Average \\
			\midrule
			nnU-Net~\cite{Isensee2021}  & 80.15$\pm$7.85 & \textbf{93.05$\pm$2.59}  & 87.33$\pm$5.84 & 93.78$\pm$1.67 & 88.57$\pm$3.84 & 3.17$\pm$5.45 & 5.79$\pm$9.38& 7.41$\pm$10.66 & 1.530$\pm$2.16 & 4.47$\pm$5.38 \\
			TransUNet~\cite{Chen2021a} &   75.66$\pm$9.36 & 85.27$\pm$16.18 & \textbf{89.37$\pm$4.26} & 91.52$\pm$2.97 & 85.46$\pm$7.20 & 2.06$\pm$1.28  & 5.15$\pm$4.60  & \textbf{1.48$\pm$0.87} & 2.93$\pm$2.26 & 2.91$\pm$1.92 \\
			nnFormer~\cite{Zhou2021a}&   78.00$\pm$6.86 & 92.47$\pm$2.06   & 84.87$\pm$10.10  & 90.07$\pm$4.89 &  86.35$\pm$3.78 & 2.76$\pm$2.76  & 5.51$\pm$4.48  & 
			6.56$\pm$16.39  & 3.07$\pm$2.59 & 4.47$\pm$4.63 \\
			UNETR++~\cite{Shaker2022} & 76.24$\pm$10.43  & 92.00$\pm$4.79  & 88.61$\pm$4.49  & 92.48$\pm$ 2.32 & 87.33$\pm$3.67 & 1.90$\pm$ 0.99 & 3.84$\pm$3.93 & 2.65$\pm$3.58& 1.19$\pm$0.56 & 2.39$\pm$1.44\\
			PCT-Net & 82.08$\pm$6.19  & 88.47$\pm$11.18  & 88.11$\pm$4.43  & 91.65$\pm$3.67 & 87.58$\pm$4.39 & 1.59$\pm$1.01 & 3.40$\pm$2.75 & 1.70$\pm$0.95& 1.26$\pm$0.66 & 1.99$\pm$1.01\\
			PCT-Net + VF & \textbf{83.45$\pm$4.78}  & 91.66$\pm$7.14   & 89.26$\pm$4.47   & \textbf{93.88$\pm$1.79} & \textbf{89.56$\pm$2.81} & \textbf{1.24$\pm$0.33} & \textbf{2.93$\pm$2.52} & 1.49$\pm$1.03 & \textbf{0.98$\pm$0.51} & \textbf{1.66$\pm$0.86} \\
			\bottomrule
		\end{tabular}%
	}
	\label{tab:sota_segthor}%
\end{table*}%

\begin{table*}
	\centering
	\caption{ Comparison of Dice (\%) obtained by different methods for abdominal organ segmentation on Synapse dataset. `R' and `L' represent right and left, respectively. Bold font and underline represent the first and second top method in each column, respectively. }
	\resizebox{\linewidth}{!}
	{
		\begin{tabular}{l|lllllllll}
			\toprule
		  Method	& Spleen  & R Kidney &   L Kidney & Gallbladder & Pancreas & Liver & Stomach  &  Aorta & Average \\
			\midrule
		 nnU-Net~\cite{Isensee2021}  & \textbf{94.00$\pm$4.26}  & 91.89$\pm$7.72  & \underline{93.30$\pm$4.17} & 78.17$\pm$18.48 & \textbf{83.27$\pm$3.98} & 94.33$\pm$3.89 & 79.30$\pm$19.89 & 89.26$\pm$3.38 &87.94$\pm$5.26 \\
		TransUNet~\cite{Chen2021a}& 92.00$\pm$7.15   & 92.48$\pm$4.11 & 92.30$\pm$4.77 & 74.21$\pm$12.31 & 72.18$\pm$16.12 & 94.73$\pm$4.11  & 75.72$\pm$15.54 & \underline{90.67$\pm$4.16} & 85.54$\pm$5.40 \\
		nnFormer~\cite{Zhou2021a}  & \underline{92.25$\pm$5.83}   & 92.86$\pm$2.11   & \textbf{93.84$\pm$1.45}  & 73.56$\pm$14.48 &  72.02$\pm$6.22 & 95.31$\pm$1.28   & 80.77$\pm$10.29 & 90.47$\pm$3.54 & 86.39$\pm$2.81 \\
		UNETR++~\cite{Shaker2022} & 89.26$\pm$15.54  & 93.40$\pm$1.61  & 93.19$\pm$2.31  & 70.96$\pm$ 28.24 & 74.70$\pm$12.14 & 95.76$\pm$ 0.68 & \textbf{82.79$\pm$15.22} & 88.79$\pm$4..82 & 86.11$\pm$6.55  \\
		PCT-Net & 91.36$\pm$13.77   & \underline{95.21$\pm$5.46}   & 90.78$\pm$9.79   & \textbf{80.94$\pm$9.91} & 79.13$\pm$9.86 & \underline{96.63$\pm$7.04} & 79.25$\pm$23.33 & 90.48$\pm$5.11 & \underline{87.97$\pm$5.22} \\
		PCT-Net + VF &   91.38$\pm$12.97   & \textbf{95.31$\pm$0.55}   & 92.17$\pm$8.00 & \underline{80.79$\pm$13.58} & \underline{83.24$\pm$3.97} & \textbf{96.70$\pm$6.30} &
		\underline{82.46$\pm$15.99} & \textbf{90.86$\pm$4.10} & \textbf{89.11$\pm$4.43}   \\
			\bottomrule
		\end{tabular}%
	}
	\label{tab:sota_synapse_dice}%
\end{table*}%

\begin{table*}
	\centering
	\caption{Comparison of ASSD (mm) obtained by different methods for abdominal organ segmentation on Synapse dataset.}
	\resizebox{\linewidth}{!}
	{
		\begin{tabular}{l|lllllllll}
			\toprule
			Method	& Spleen  & R Kidney &   L Kidney & Gallbladder & Pancreas & Liver & Stomach  &  Aorta & Average \\
			\midrule
			nnU-Net~\cite{Isensee2021}  & 4.82$\pm$7.90  & 3.56$\pm$8.00  & 4.96$\pm$9.78 & 2.32$\pm$2.57 & 1.92$\pm$1.24 & 3.11$\pm$4.39 & 6.43$\pm$11.02 & 3.42$\pm$3.36 & 3.82$\pm$5.33 \\
			TransUNet~\cite{Chen2021a}& 6.34$\pm$11.64   & 2.69$\pm$3.77 & 4.20$\pm$7.37 & 4.73$\pm$6.31 & 3.55$\pm$1.69 & 2.17$\pm$2.16  & 6.65$\pm$8.28 & 2.91$\pm$3.19 & 4.15$\pm$5.16 \\
			nnFormer~\cite{Zhou2021a}  & 6.07$\pm$9.21   & 1.70$\pm$2.24   & \textbf{1.25$\pm$1.61}  & 6.63$\pm$9.88 &  4.35$\pm$3.00 & 2.27$\pm$2.29   & 7.84$\pm$13.15 & 1.88$\pm$1.89 & 4.00$\pm$5.04 \\
			UNETR++~\cite{Shaker2022} & 4.87$\pm$10.34  & 1.63$\pm$1.90  & 1.87$\pm$3.24  & 3.22$\pm$ 4.76 & 3.69$\pm$3.92 & 1.46$\pm$ 0.67 & 4.65$\pm$6.20 & 1.74$\pm$0.84 & 2.89$\pm$3.31  \\
			PCT-Net & 4.07$\pm$9.63   & 0.50$\pm$0.07  & 3.10$\pm$6.77   & 3.20$\pm$3.78 & 3.06$\pm$3.98 & 1.04$\pm$0.34 & 5.35$\pm$8.45 & \textbf{1.55$\pm$1.02} & 2.73$\pm$3.00 \\
			PCT-Net + VF &   \textbf{4.04$\pm$9.20}   & \textbf{0.47$\pm$0.06}   & 2.05$\pm$4.22 & \textbf{1.77$\pm$1.98} & \textbf{1.52$\pm$0.65} & \textbf{1.01$\pm$0.29} &
			\textbf{3.64$\pm$3.81} & 1.85$\pm$2.06 & \textbf{2.04$\pm$2.09}   \\
			\bottomrule
		\end{tabular}%
	}
	\label{tab:sota_synapse_assd}%
\end{table*}%

\textbf{Comparison of Different Models.} 
Our PCT-Net pretrained with Volume Fusion (VF) on PData-110k was compared with four state-of-the-art segmentation models: nnU-Net~\cite{Isensee2021}, TransUNet~\cite{Chen2021a}, UNETR++~\cite{Shaker2022} and nnFormer~\cite{Zhou2021a}. They were evaluated on MICCAI 2015 Head-Neck dataset, SegTHOR and Synapse datasets, respectively.

Table~\ref{tab:sota_hn} shows quantitative evaluation results on the MICCAI 2015 Head-Neck dataset. The nnU-Net~\cite{Isensee2021} achieved an average Dice of 78.66\%, with the mandible and optic chiasm being the easiest and hardest OARs to segment, respectively. TransUNet and nnFormer failed to beat nnU-Net, with average Dice of  69.10\% and 70.27\%, respectively. UNETR++ outperformed nnU-Net with average Dice of 80.37\%. Our PCT-Net achieved an average Dice of 81.41\%, and performed better than the existing networks. When PCT-Net was pretrained with Volume Fusion, the average Dice was further improved to 82.74\%. The average ASSD obtained by PCT-Net + VF was 0.77~mm, which was better than those of the other methods. Fig.~\ref{fig:oar_miccai15} shows a visual comparison between the segmentation results obtained by different methods. It can be observed that TransUNet and nnFormer have a poor performance on tiny structures such as optic chiasm and optic nerves. The nnFormer also has an over-segmentation of the brain stem in the first row. UNETR++ has an over-segmentation in the mandible, as shown in the second row. PCT-Net has an overall better performance than these methods, and when pretrained with VF, its performance is further improved, especially for optic nerves as shown in the first row.  

Table~\ref{tab:sota_segthor} shows quantitative evaluation of different methods for thoracic OAR segmentation on SegTHOR dataset. Among the existing methods, nnU-Net achieved the highest average Dice (88.57\%). However, the average ASSD value (4.47~mm) obtained by nnU-Net was lower than that of TransUNet (2.91~mm) and UNETR++ (2.39~mm). PCT-Net obtained an average Dice of 87.58\% and ASSD of 1.99~mm, respectively. When pretrained with VF, the corresponding Dice and ASSD values were 89.56\% and 1.66~mm, respectively, which outperformed the other methods.

\begin{figure*}[t]
	\centering
	\includegraphics[width=1.0\linewidth]{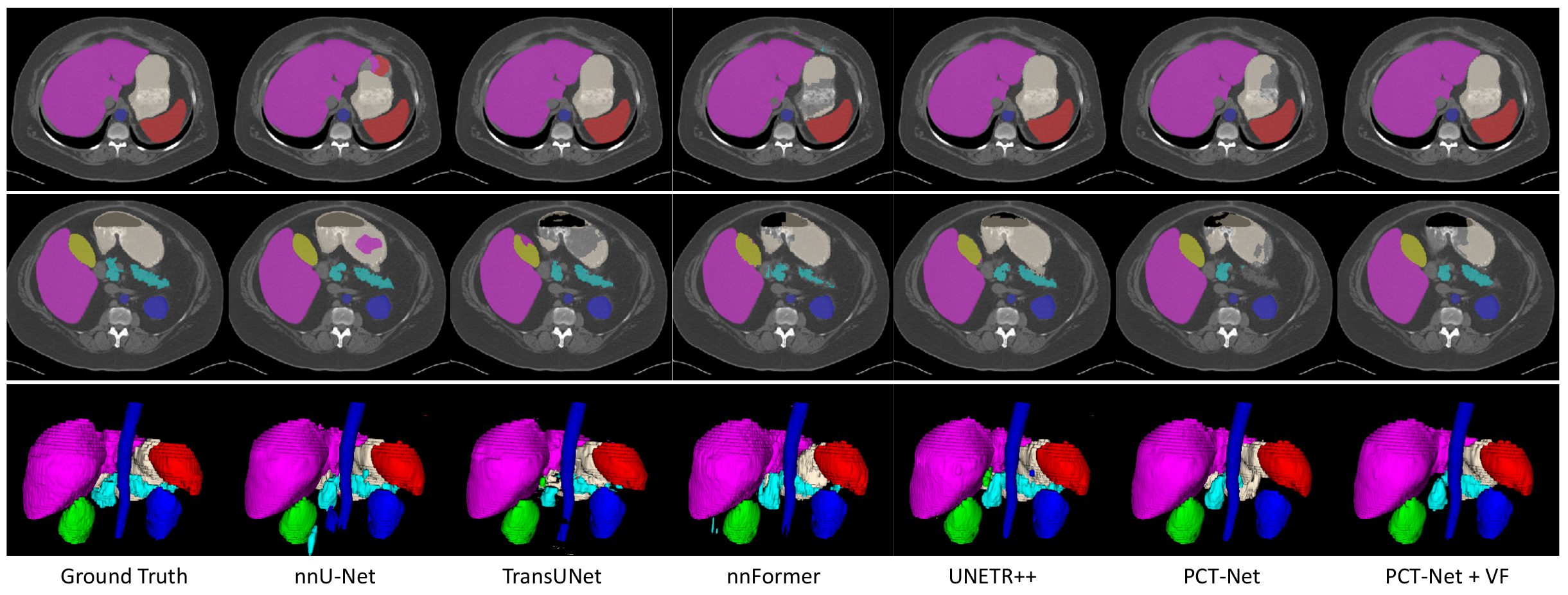}
	
	\caption{Visual comparison between results of different methods on Synapse dataset for abdominal organ segmentation. The first two rows show results in two different axial slices, and the last row presents 3D visualization.}
	\label{fig:synapse}
\end{figure*}

Table~\ref{tab:sota_synapse_dice} and \ref{tab:sota_synapse_assd} show the quantitative evaluation of different methods on the Synapse dataset in terms of Dice and ASSD, respectively.  The average Dice of nnU-Net was 87.94\%, which was better than that of TransUNet (85.54\%), nnFormer (86.39\%) and UNETR++ (86.11\%). The average Dice of PCT-Net (87.97\%) was slightly higher than that of nnU-Net, and its average ASSD value (2.73~mm) was much better than that of nnU-Net (3.82~mm). When pretrained with VF, the performance of PCT-Net was further improved, with average Dice and ASSD of 89.11\% and 2.04~mm, respectively.  Note that nnU-Net was a strong segmentation framework. Despite that it is hard to defeat nnU-Net on all the organs, our method outperformed  nnU-Net on 5 out of the 8 organs in terms of Dice, and on all the 8 organs in terms of ASSD. Fig.~\ref{fig:synapse} presents a visual comparison between the segmentation results of these different methods, which shows that PCT-Net + VF achieved the best performance overall compared with the other methods. 

\section{Conclusion}
In this work, we present a self-supervised foundation model pretrained with large-scale unannotated volumetric images for 3D medical image segmentation. A volume Fusion method is proposed for self-supervised pretraining, where the pretext task is formulated as a pseudo-segmentation task by forcing the model to predict the discrete fusion coefficient of each voxel between two different sub-volumes. It encourages the model to learn multi-scale contextual information that is transferable to downstream segmentation tasks. A novel Parallel Convolution and Transformer Network (PCT-Net) is also proposed for better feature representation ability in 3D segmentation tasks.  Our PCT-Net was pretrained with Volume Fusion on 110k unannotated CT scans, and experimental results  showed that our pretrained model outperformed several state-of-the-art segmentation methods on different downstream tasks for multi-organ segmentation from CT images. In the future, it is of interest to validate our pretrained model on downstream tasks for lesion and organ segmentation from different modalities.

\section{Acknowledgment}
This work was supported by the National Natural Science Foundation
of China (62271115), National Key Research and Development
Program of China (2020YFB1711500) and the 1·3·5 project
for disciplines of excellence, West China Hospital, Sichuan University
(ZYYC21004).

{\small
\bibliographystyle{ieee_fullname}
\bibliography{D:/Documents/texstudio/latex_reference/self_sup}
}

\end{document}